# Enhanced MRI brain tumor detection using deep learning in conjunction with explainable AI SHAP based diverse and multi feature analysis

Asif Rahman[1], Maqsood Hayat[1]✉, Nadeem Iqbal[1], Fawaz Khaled Alarfaj[2], Salem Alkhalaf[3]✉ & Fahad Alturise[4]

**Recent innovations in medical imaging have markedly improved brain tumor identification, surpassing conventional diagnostic approaches that suffer from low resolution, radiation exposure, and limited contrast. Magnetic Resonance Imaging (MRI) is pivotal in precise and accurate tumor characterization owing to its high-resolution, non-invasive nature. This study investigates the synergy among multiple feature representation schemes such as local Binary Patterns (LBP), Gabor filters, Discrete Wavelet Transform, Fast Fourier Transform, Convolutional Neural Networks (CNN), and Gray-Level Run Length Matrix alongside five learning algorithms namely: k-nearest Neighbor, Random Forest, Support Vector Classifier (SVC), and probabilistic neural network (PNN), and CNN. Empirical findings indicate that LBP in conjunction with SVC and CNN obtained high specificity and accuracy, rendering it a promising method for MRI-based tumor diagnosis. Further to investigate the contribution of LBP, Statistical analysis chi-square and p-value tests are used to confirm the significant impact of LBP feature space for identification of brain Tumor. In addition, The SHAP analysis was used to identify the most important features in classification. In a small dataset, CNN obtained 97.8% accuracy while SVC yielded 98.06% accuracy. In subsequent analysis, a large benchmark dataset is also utilized to evaluate the performance of learning algorithms in order to investigate the generalization power of the proposed model. CNN achieves the highest accuracy of 98.9%, followed by SVC at 96.7%. These results highlight CNN's effectiveness in automated, high-precision tumor diagnosis. This achievement is ascribed with MRI–based feature extraction by combining high resolution, non-invasive imaging capabilities with the powerful analytical abilities of CNN. CNN demonstrates superiority in medical imaging owing to its ability to learn intricate spatial patterns and generalize effectively. This interaction enhances the accuracy, speed, and consistency of brain tumor detection, ultimately leading to better patient outcomes and more efficient healthcare delivery. https://github.com/asifrahman557/BrainTumorDetection.**

The human brain, a fundamental component of the central nervous system (CNS), collaborates with the spinal cord to manage and synchronize vital body activities, such as decision-making, signal transmission, and sensory information integration[1]. The complicated and highly structured anatomy facilitates complex cognitive and physiological processes; nevertheless, this complexity also makes it susceptible to various illnesses and diseases, including stroke, infections, and brain tumors[2,3]. Brain tumors, characterized by the unregulated growth of abnormal cells within the limited confines of the skull, present considerable threats to neurological function and general health[4–6]. These tumors are generally categorized into two types: benign (non-cancerous) and malignant

[1]Department of Computer Science, Abdul Wali Khan University Mardan, Mardan, Khyber-Pakhtunkhwa, Pakistan. [2]Department of Management Information Systems (MIS), School of Business, King Faisal University (KFU), 31982 Al-Ahsa, Saudi Arabia. [3]Department of Computer Engineering, College of Computer, Qassim University, Buraydah, Saudi Arabia. [4]Department of Cybersecurity, College of Computer, Qassim University, Buraydah, Saudi Arabia. ✉email: m.hayat@awkum.edu.pk; s.alkhalaf@qu.edu.sa

(cancerous), with more than 200 specific subtypes recognized based on their cellular origin, location, and behavior[7,8]. Despite the advancement in medical research, the prognosis for patients with brain tumors remains unfavorable, with survival rates differing markedly based on tumor type, location, and stage at diagnosis[9–11].

Brain tumors are a significant source of morbidity and mortality globally, ranking as the 10th most prevalent cause of death across all age demographics[12]. The National Brain Tumor Foundation (NBTF) reports a 300% increase in brain tumor-related fatalities over the last three decades, highlighting the critical necessity for enhanced diagnostic and treatment approaches. Timely identification and care are essential, since untreated tumors may result in significant neurological impairments or death[13]. Conventional diagnostic techniques, including biopsies, are intrusive, protracted, and fraught with hazards, hence demanding the creation of non-invasive, precise, and efficient diagnostic instruments. Magnetic Resonance Imaging (MRI) has emerged as the benchmark for diagnosing brain tumors, providing exceptional visibility of analytical structures and anomalies.

Brain tumors, originating inside the brain, constitute 14.1% of yearly occurrences, with a disproportionately elevated frequency in children (70% of cases)[14,15]. Despite their commonality, efficacious therapies for several brain tumor subtypes continue to be elusive, and the enduring adverse consequences of these tumors on patients' quality of life are significant[16,17]. Moreover, global statistics reveal a significant rise in brain tumor manifestations, with incidence rates escalating by 10–15% from 2004 to 2020[18]. These numbers emphasize the increasing prevalence of brain tumors and the urgent inevitability of novel approaches to enhance detection, treatment, and patient outcomes. However, the prompt and efficient care of brain cancers depends on precise diagnosis and categorization. The variability of brain tumors regarding size, location, form, and imaging properties, together with changes in scanning settings and modalities, renders this endeavor intrinsically hard[19]. Conventional diagnostic methods, such as the Leksell Gamma Knife and radioactive beams, while beneficial, are labor-intensive, need considerable human knowledge, and frequently entail extended processing durations[20]. The emergence of sophisticated imaging technologies, such as Computer Tomography (CT), MRI, and Positron Emission Tomography (PET), has transformed brain tumor diagnosis by offering intricate, non-invasive viewing of cerebral structures. MRI is the preferred modality because of its higher soft tissue contrast and capability to acquire multi-parametric data, including T1-weighted, T2-weighted, and FLAIR sequences[21].

Nevertheless, with these breakthroughs, the intricacy of brain tumors persists in posing considerable obstacles for doctors and researchers. The necessity for accurate, automated diagnostic instruments is essential, as early identification and commencement of therapy are pivotal elements affecting patient survival rates[22]. The amalgamation of sophisticated imaging data with machine learning (ML) and deep learning (DL) methodologies presents a potential approach to overcoming these problems, facilitating the creation of intelligent systems proficient in precise tumor identification, classification, and segmentation.

The rapid advancement of machine learning (ML) and deep learning (DL) has revolutionized medical imaging, providing robust instruments for the automated examination of intricate datasets[23,24]. Deep Neural Networks (DNNs), especially Convolutional Neural Networks (CNNs), have exhibited exceptional efficacy in image classification, object recognition, and natural language processing applications[25–26]. These breakthroughs are driven by enhancements in computing power, particularly GPUs, and the accessibility of extensive, annotated datasets, which have enabled the creation of very precise models for medical applications[27,28].

In the realm of brain tumor diagnostics, deep learning models have demonstrated remarkable potential. CNNs, due to their hierarchical structure and capacity to learn spatial feature hierarchies, are especially adept in analyzing medical pictures[29]. Techniques including autoencoders, transfer learning, and ensemble approaches have significantly improved the efficacy of these models, allowing them to attain state-of-the-art outcomes in tumor detection and classification[30–32]. Hybrid models that integrate CNNs with optimization methods, like Particle Swarm Optimization (PSO) and Genetic methods (GAs), have exhibited enhanced efficacy in tumor segmentation and classification tasks[33,34].

Despite these achievements, other difficulties persist. Numerous current investigations are constrained by the utilization of small, homogenous datasets, which may insufficiently reflect the range of tumor types, imaging modalities, and patient demographics[35,36]. Furthermore, the incorporation of deep learning models into clinical practice necessitates the resolution of challenges of computational complexity, interpretability, and generalizability[37]. Moreover, the absence of common assessment measures and benchmarking datasets complicates the comparison of various models and methodologies[38].

Substantial advancements have been achieved in the formulation of computer models for brain tumor diagnosis, with an increasing focus on intelligent methodologies. Recent research has shown the capability of machine learning and deep learning approaches to revolutionize brain tumor diagnostics. Almadhoun et al.[39] attained 98% test accuracy utilizing a deep learning model on a dataset including 10,000 MR images. Musallam et al.[40] suggested a DCNN model achieving an overall accuracy of 97.72% for the classification of four tumor subtypes. Additional significant contributions encompass the use of correlation learning techniques[41], ensemble classifiers[42], and sophisticated designs such as Dense EfficientNet[43], which attained accuracies beyond 98%.

Hybrid models, like those that integrate CNNs with SVM or evolutionary algorithms, have demonstrated encouraging outcomes. Mittal et al.[44] shown enhanced segmentation accuracy by the integration of Growing CNN with Stationary Wavelet Transform. Preprocessing approaches, including picture augmentation and multimodal information fusion, have significantly improved the performance of these models[45,46]. In addition, recent studies in unsupervised domain adaptation have demonstrated promising performance in medical image segmentation such as multi-organ segmentation[47], knee MRI analysis[48], identification of ocular diseases[49]. Furthermore, liver segmentation across modalities has been approached using adversarial and self-learning techniques[50].

In a sequel, ML approaches tackle more difficulties associated with dataset variety and computational complexity[51,52]. The detailed analysis of the literature review is illustrated in Table 1.

| Classification | Reference | Method/model | Details | Performance metrics |
|---|---|---|---|---|
| Deep learning models | Almadhoun et al.[53]. , | Educational Model utilizing Deep learning algorithms | Transfer learning<br>10,000 MRI images | Training accuracy: 100%<br>Testing accuracy: 98% |
| | Musallam et al.,[54] | DCCN model | Multi categories tumor classification | Accuracy: 97.72% |
| | Nayat et al.,[55] | Dense efficient/Net | Hybrid Model | Accuracy: 98.78% |
| | Obeidavi et al.,[56] | Residual Network based on CNN | BRATS dataset 2015 | Accuracy: 97.05% |
| | Sajjad et al.,[57] | Hybrid CNN model | CNN + Additional approaches | Accuracy: 86%<br>Specificity: 91% |
| | Paual et al.[52] | CNN and Fully connected network | 3064 MRI images | Accuracy: 91.43% |
| | Mittal et al.,[60] | Stationary wavelet Transform and CNN | Focus on segmentation | -- |
| | Cinar and Yildirim[63] | GoogLeNet, InceptionV3, DenseNet201, AlexNet, and ResNet-50 | Tumor segmentation and localization | |
| | Sangeeta et al.,[66] | ResNet-50 | Brain Tumor classification | Accuracy: 96.50% |
| | Liang et al.,[65] | Fusion of DL model | Brain tumor scaling | |
| Machine learning algorithms | Wozniak et al.[58] | Correlation learning model | 3064 brain cancer images | Accuracy: 96% |
| | Garg et al.,[59] | Hybrid model | | Accuracy: 97.3% |
| | Lotlikar et al.,[20] | K-nearest neighbor | Brain dataset | Accuracy: 95.6% |
| | Gumaei et al.,[47] | RELM + PCA-NGIST | | Accuracy: 94.2% |
| | Pasheal et al.,[64] | CNN and 10-fold cross-validation | | Accuracy: 93.6% |
| | Narayana and Reddy[61] | Ensemble model | MRI images | Accuracy; 91% |
| Optimization based model | Dixit and Nanda[62] | PSO and CNN model | Brain tumor detection | Accuracy: 92% |

**Table 1**. Literature review.

This study aims to address the limitations of existing models or predictors by assessing the effectiveness of various machine learning and deep learning models with diverse feature extraction strategies for brain tumor classification in order to achieve high diagnostic accuracy. This research endeavors to provide a robust framework for accurate and timely brain tumor diagnosis, ultimately improving patient outcomes.

The contribution of this study includes:

- A comprehensive assessment of cutting-edge machine learning and deep learning models for brain tumor categorization.
- The integration of sophisticated feature extraction methods to enhance model performance.
- A comparative analysis of the proposed model with existing methodologies, highlighting their limitations and strengths.

## Materials and methods

### Dataset

This research utilizes the publicly accessible Brain Tumor MRI dataset from Kaggle, https://www.kaggle.com/datasets/masoudnickparvar/brain-tumor-mri-dataset[67]. The dataset consists of MRI images grouped into two primary categories tumor and non-tumor. Further tumor images were categorized into three subtypes including Pituitary Tumors, Meningioma tumors, and Glioma Tumors. The dataset consists of 2000 MRI images of which 1000 images of non-tumor and 1000 images of tumor. Further, tumor data is composed of 300 images of Pituitary, 400 of Meningioma, and 300 images of Glioma tumors. In order to evaluate the generalization of the model, a large training dataset is used, which consists of 7023 MRI images split into 5712 images for training and 1311 images for testing[63]. In the training dataset, 4117 images are affected by tumors further classified into 1457 images of the pituitary, 1339 images of meningioma, and 1321 images of glioma tumors whereas 1595 images of non-tumor. The testing dataset consists of 1311 images of which 405 images of non-tumors and 906 images of tumors. Tumor images are 300 images of pituitary, 306 images of meningioma, and 300 images of glioma tumors.

This work employed a series of preprocessing and data cleansing steps to augment the quality of brain tumor MRI images and boost the precision and resilience of the classification models. Initially, all images were transformed to grayscale to minimize computational complexity by removing superfluous color channels, as the essential characteristics for classification are based on intensity fluctuations rather than color information. A Gaussian filter was subsequently added to blur images, therefore, diminishing high-frequency noise and enabling the model to concentrate on the most pertinent characteristics while lessening the impact of slight alterations[68]. Subsequently, the binary threshold was executed to convert the grayscale images into binary images. During this process, the pixels' value above the predefined threshold were rendered white (foreground) and those below were assigned black (background). This stage highlighted the delineation of objects, especially tumor areas, thereby facilitating the distinction between malignant tissue and healthy tissue. In a sequel, contour detection methods were utilized to ascertain the greatest contour, which generally represents the tumor region, in each image. The images were subsequently cropped to emphasize the area of interest (ROI), eliminating extraneous background elements and ensuring that the model focused on the most relevant aspects of the image. Ultimately, all images were downsized to a uniform resolution of 128 × 128 pixels to maintain uniformity throughout the collection.

Further, several advanced data augmentation techniques were utilized to enhance the diversity of the benchmark training dataset and strengthen the model's ability to generalize to novel data. Data augmentation is a technique that artificially increases the quality such as illumination, size, and orientation as well as the diversity of a dataset by applying various transformations. This research utilized random modifications to brightness and contrast. Brightness was arbitrarily adjusted within a range of 0.8 to 1.2, emulating diverse lighting situations that may arise during image acquisition. Likewise, the contrast was modified within the identical range to assist the model in identifying tumor characteristics across varying imaging settings. Subsequently, all enhanced images were normalized by dividing the pixel values by 255, so scaling them to a range between 0 and 1.

## Feature representation techniques

This research examined six distinct feature representation strategies that are essential for transforming MRI images into significant data, which is subsequently employed for classification. The framework of proposed model is illustrated in Fig. 1. Comprehensive details of these feature representation approaches are listed below:

### *Convolutional neural networks (CNN)*

CNNs are a type of deep learning algorithms that automatically learn both discriminatory and hierarchical information from input data[69]. Here, CNNs were utilized to extract information from brain MRI data. The pre-trained CNN architecture proved notably proficient in capturing intricate spatial patterns, textures, and structural details, which contributed to obtaining high throughput for brain tumor detection. CNN can be formally expressed as:

$$Feature\ space\ (x+y) = \sum{}_{i=1}^{k} \sum{}_{j=0=1}^{k} input\,(x+i,\ y+j)\,.\,Kernal\ (i,\ j) \tag{1}$$

Where $input\,(x+i,\ y+j)$ represents the input image pixel values, $Kernal\ (i,\ j)$ denotes the filter weights, k is the size of the kernel.

The hierarchical discriminative attributes extraction of CNNs facilitated the recognition of elusive motifs revealing of brain tumors.

### *Discrete wavelet transform (DWT)*

The Discrete Wavelet Transform (DWT) is a signal processing method that expresses both time and frequency analysis simultaneously. Here, DWT was utilized on MRI images to split into frequency sub-bands in order to extract both spatial and frequency information. The wavelet transform can be formulated as:

$$W\ (x,\ y) = \frac{1}{\sqrt{x}} \int_{-\infty}^{\infty} image\ (t) + \ \psi\ \left(\frac{t-y}{x}\right) dt \tag{2}$$

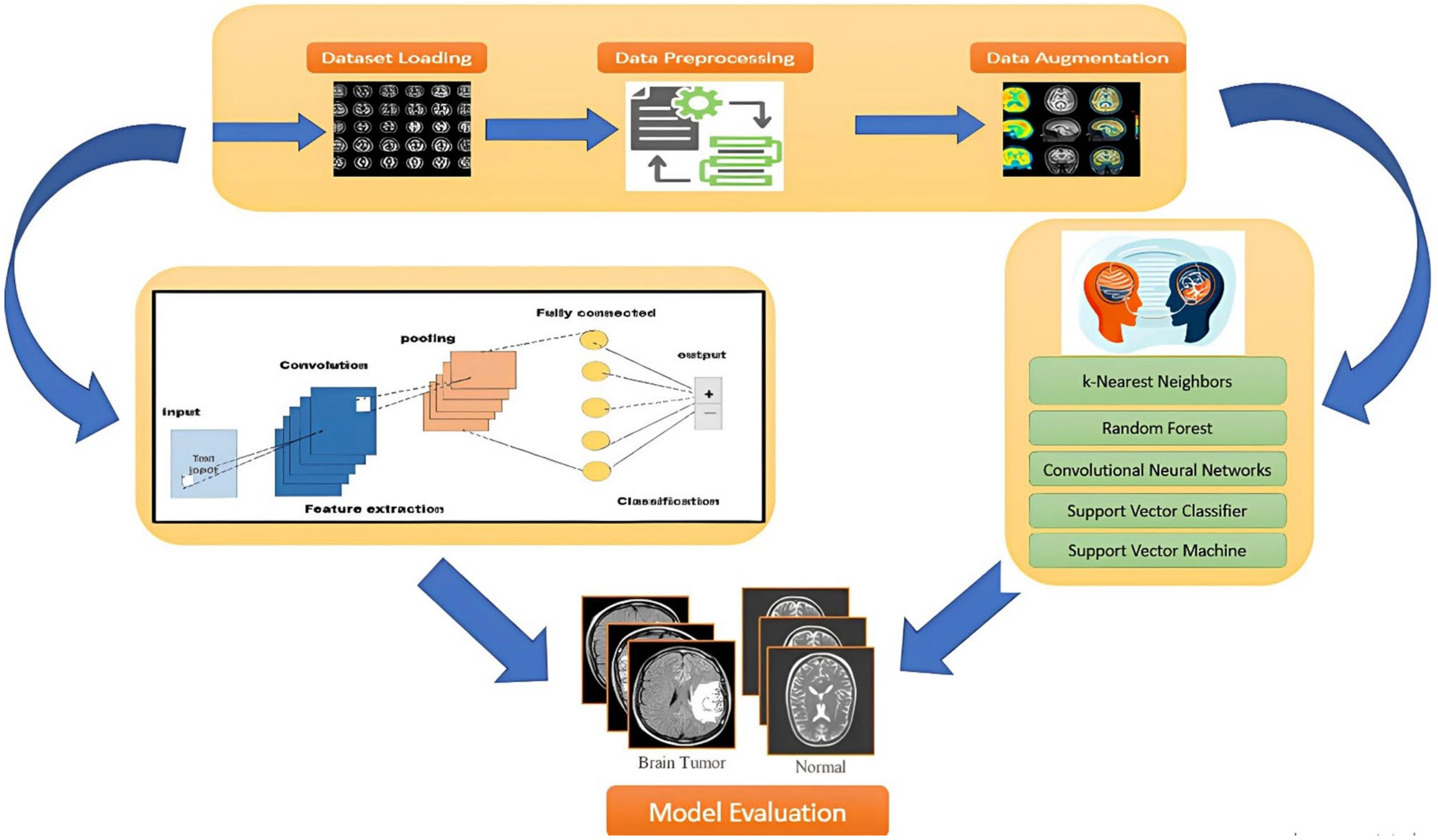

**Fig. 1**. Framework of proposed model.

Where $x$ is the scaling factor, y is the translation factor, and $\psi(t)$ is the wavelet function.

The outcome of DWT yielded a comprehensive representation of the image information which assist in comprehension of spatial and frequency characteristics of MRI images.

*Fast fourier transform (FFT)*

The Fast Fourier Transform (FFT) is a mathematical tool used to transform a signal from the spatial domain frequency domain. Here FFT was utilized to extract frequency-based information from MRI images. The FFT of an image Input(x, y) is given by:

$$F\ (u,v)=\sum\ _{x=1}^{M-1}\sum_{y=1}^{N-1}Input\,(x,\ y)\,.\,e^{-i2\pi\,(\frac{ux}{M}+\frac{vy}{N})} \tag{3}$$

where $F(u,v)$ denotes the frequency domain representation, $M$ and $N$ are the dimensions of the image. FFT-based frequency-domain information highlights patterns and structural information which is useful in distinguishing between healthy and tumor-affected brain images.

*Gabor filters*

Gabor Filters are effective tools which is used for the extraction of texture features from images. These filters are inspired by the human visual system and are mostly effective in collecting local spatial and frequency attributes The 2D Gabor filter is defined as:

$$G\,(x,\ y)=exp\left(\frac{x'^{\,2}+\gamma^{\,2}y'^{\,2}}{2\delta^{\,2}}\right).\cos\left(2\pi\,\frac{x'}{\lambda}+\psi\right) \tag{4}$$

$$x' = x\cos + y\sin,\ y' = x\sin + y\cos,$$

.

where $\gamma$ is the spatial aspect ratio, $\lambda$ is the wavelength, $\theta$ is the orientation, $\sigma$ is the standard deviation, $\psi$ is the phase offset. Here Gabor Filters were applied to capture texture attributes from MRI images, which assisted in detecting patterns revealing brain tumors.

*Gray-level run length matrix (GLRLM)*

The Gray-Level Run Length Matrix (GLRLM) is a texture analysis method that measures the frequency of contiguous pixels with the same gray level value. This approach was utilized to extract texture information from MRI data. The GLRLM is defined as:

$$P\,(i,\ j|\theta\,)=Number\ of\ runs\ with\ gray\ level\ i\ and\ run\ length\ j\ in\ direction\ \theta \tag{5}$$

where: $i$ is the gray level, $j$ is the run length, $\theta$ is the direction. The retrieved texture attributes facilitate the distinction between tumor-affected and healthy brain tissues by detecting textural variations in the MRI images.

*Local binary patterns (LBP)*

Local Binary Patterns (LBP) is a texture-oriented feature extraction method that evaluates each pixel with its adjacent pixels to generate a binary pattern. The LBP value for a pixel ($x_c$,$y_c$) is determined as follows:

$$LBP\,(x_c,\ y_c)=\sum_{p=0}^{p-1}\mathrm{s}\,(g_p-g_c)\,2^p \tag{6}$$

where: $gc$ is the gray value of the center pixel, $gp$ is the gray value of the adjacent pixel, $P$ is the number of neighbors, $s(x)$ is a threshold function defined as:

$$\mathrm{s}\,(x)=\left\{\begin{array}{l}1\ if\ x\ \geq\ 0\\ 0\quad\ \ x\ <0\end{array}\right.$$

Here, LBP was applied to identify textural modification in MRI images, which assisted in recognizing tumor-affected regions.

The utilized feature representation methods: CNN, DWT, FFT, Gabor Filters, GLRLM, and LBP are pivotal in transforming raw MRI images into significant data for classification. Each method played promising role in identifying distinct attributes, such as spatial patterns, frequency data, and texture properties, which help the learning algorithms to detect brain tumors accurately and effective.

## Evaluation metrics and performance measures

Here in this section, it is thoroughly defined how well the various feature representation techniques worked with different machine learning models for categorizing the MRI scan images. The objective is to know how these models with a combination of various feature representation techniques worked to distinguish the benign and malignant brain images. To accomplish this task, accuracy, F1-score, precision, sensitivity, and specificity were included in the set of defined evaluation criteria. Every metric was computed to evaluate the result of these classification techniques. These five basic evaluation metrics are defined as under which were used in series of publication[70–74]. In addition, five-fold stratified cross-validation test was adopted to ensure fair evaluation and address class imbalance. In five-fold cross validation test, data was split into five folds such that each fold preserved the same class distribution as the overall dataset. This method ensures that every tumor subtypes, including minority classes are proportionally represented in each training and validation test, this technique enhances the reliability of performance metrics.

Accuracy:

The overall correctness of a classifier can be evaluated by these metrics. It is considered the most effective aspect to be measured. The accuracy is actually when we divide the number of correct predictions by the total number of predictions. The formula for this is given as under.

$$Accuracy = \frac{True\ Positives + True\ Negatives}{False\ Positives + False\ Negatives + True\ Positives + True\ Negatives} \tag{7}$$

Sensitivity (recall or true positive rate):

The ratio of true positive cases and true positives plus false negatives is called the sensitivity of the model. It can be calculated using the following formula.

$$Sensitivity = \frac{True\ Positives}{False\ Negatives + True\ Positives} \tag{8}$$

Precision:

The ratio between true positives and true positives plus false positives is known as the precision of the classifier. The formula of precision is given below.

$$Precision = \frac{True\ Positives}{False\ Positives + True\ Positives} \tag{9}$$

F1 score:

The F1 score is also a very important measure of the efficiency of a model. It can be calculated by using the precision and sensitivity as given.

$$F1\ Score = \frac{2\ *\ Precision\ *\ Sensitivity}{Precision + Sensitivity} \tag{10}$$

Specificity (true negative rate):

The specificity of a model determines how a model distinguishes between all the actual negatives and the calculated negatives. The formula is given as under.

$$Specificity = \frac{True\ Negatives}{True\ Negatives + False\ Positives} \tag{11}$$

- True Positives are those cases that are correctly predicted as positive.
- False Positives are those cases that are erroneously predicted as positive.
- True Negatives are those cases that are correctly predicted as negative.
- False Negatives are those cases that are erroneously predicted as negative.

## Result analysis and discussion

This section discusses comprehensively the performance assessment of various feature representation techniques (CNN, DWT, FFT, Gabor, GLRLM, and LBP) in conjunction with several machine learning and deep learning algorithms. Performance Analysis of Feature Extraction Techniques with ML Classifiers. Class weighting is implemented during training to rectify the class imbalance in the sizable dataset, giving the minority (non-tumor) class larger loss penalties. Furthermore, we reported additional performance metrics that better reflect performance on imbalanced datasets, such as precision and F1-score, in order to ensure a fair evaluation.

### CNN feature space

CNN serves as both a feature extraction and classification technique in this work. To extract deep features from the MRI images, we used a pre-trained CNN model (such as imageNet in the manuscript) and frozen the convolutional layers during inference. For the final classification, these features were subsequently fed into different machine learning classifiers (such SVM, KNN, or PNN). The CNN was not fine-tuned or trained end-to-end with the classifier. This fixed-feature extraction approach was chosen to reduce training time and computational cost, and to evaluate the effectiveness of transfer learning in combination with classical classifiers.

In case of classification, a CNN is employed in an end-to-end fashion, where the network simultaneously learns feature representations and performs classification. The convolutional layers act as learnable feature extractors, and the resulting features are passed through fully connected layers to generate the final classification. All layers are trained jointly from scratch without using pre-trained weights.

The CNN-based image representation approach demonstrated remarkable performance across all learning algorithms are reported in Table 2. Among these algorithms, the SVC algorithm outperformed compared to other algorithms, and achieved 96.6% accuracy and 98% specificity, highlighting its effectiveness in identifying MRI-based tumors. So the success rates of SVC are balanced in the case of all used metrics. On the other hand, RF and CNN exhibited high precision and specificity ensuring reliable positive predictions. In contrast, the sensitivity of RF and CNN is unbalanced. So the models are biased toward one class. kNN obtained 81% accuracy and 80% F1-score, while sustaining success rates. While PNN performed moderately achieved 83% accuracy and 88% specificity. In this study, robust features such as spatial hierarchies in images are captured due to CNN capabilities and eradicating the requirements of manual feature engineering. In contrast, it is computationally expensive owing to its complex structure and also shows that it is not optimal for distance-based learning. In CNN the learned attributes are not interpretable, so it is sometimes difficult to make a decision. It is overfit in the case of a small dataset.

| ML Models | Accuracy | Sensitivity | Precision | F1 Score | Specificity |
|---|---|---|---|---|---|
| kNN | 81% | 77% | 83% | 80% | 85% |
| RF | 92% | 85% | 100% | 92% | 100% |
| CNN | 96% | 92% | 100% | 96% | 100% |
| SVC | 96.6% | 95% | 96.7% | 96.1% | 98% |
| PNN | 83% | 78% | 87% | 82% | 88% |

**Table 2**. Performance evaluation metrics of various learning algorithms using CNN feature space.

| ML Models | Accuracy | Sensitivity | Precision | F1 Score | Specificity |
|---|---|---|---|---|---|
| kNN | 86% | 93% | 81% | 87% | 79% |
| RF | 90% | 81% | 100% | 89% | 100% |
| CNN | 95% | 95% | 100% | 97% | 95% |
| SVC | 95.9% | 94% | 97.0% | 95.4% | 97.8% |
| PNN | 93% | 99% | 89% | 93% | 87% |

**Table 3**. Performance evaluation metrics of various used learning algorithms utilizing DWT feature space.

| ML Models | Accuracy | Sensitivity | Precision | F1 Score | Specificity |
|---|---|---|---|---|---|
| kNN | 76% | 97% | 68% | 80% | 55% |
| RF | 83% | 91% | 79% | 84% | 76% |
| CNN | 94% | 94% | 93% | 94% | 93% |
| SVC | 95.2% | 93% | 96.44% | 95.6% | 97.4% |
| PNN | 75% | 95% | 68% | 79% | 56% |

**Table 4**. Performance evaluation metrics of various used learning algorithms utilizing FFT feature space.

## Discrete wavelet transform (DWT) feature space

Discrete Wavelet Transform (DWT) yielded robust outcomes in which SVC still led with 95.9% accuracy and 97.8%% specificity ensuring reliable predictions are shown in Table 3. In a sequel, CNN and RF also exhibited strong classification power with 100% precision. The kNN model showed excellent sensitivity of 93% and a balanced F1-score, emphasizing its capability to recognize true positives effectively. In addition, PNN also performed well with 99% sensitivity and 93% accuracy. The significance of DWT is the multi-resolution analysis which extracts both frequency and spatial information. In contrast, it requires careful consideration in the case of wavelet functions and decomposition level selection.

## FFT feature space

Fast Fourier Transform (FFT) features obtained optimal success rates using SVC, which are 97.4% specificity and over 95.2% accuracy reported in Table 4. CNN maintained a high precision rate of 93%, while kNN demonstrated effectiveness in true positive detection with is sensitivity of 97%. The sensitivity of RF is high but low specificity which reveals imbalanced classification results. FFT is computationally efficient compared to DWT along with extracting global frequency information. In contrast, it lacks spatial information, which restrains its effectiveness for localized information. FFT is sensitive to noise and assumes the signal is periodic, which may not be always true in real-time applications.

## Filters Gabor feature space

The Gabor representation method demonstrated moderated results which are shown in Table 5. In this method, SVC continuously reflected exceptional performance with 93.6% accuracy and 95.4% specificity. CNN provided a commendable precision rate of 90%, while kNN and PNN achieved comparatively lesser accuracy at 74%, indicating that Gabor features may lack of discriminative power of alternative methods. The importance of the Gabor method is the extraction of texture and edge information. On the other hand, it is computationally expensive owing to the multi-scale and multi-orientation nature of Gabor filters. Gabor filter focuses on local information while ignoring global patterns in data.

| ML Models | Accuracy | Sensitivity | Precision | F1 Score | Specificity |
|---|---|---|---|---|---|
| kNN | 74% | 67% | 78% | 72% | 81% |
| RF | 81% | 73% | 87% | 79% | 88% |
| CNN | 89% | 88% | 90% | 89% | 91% |
| SVC | 93.6% | 91% | 94.4% | 94.2% | 95.4% |
| PNN | 74% | 67% | 78% | 72% | 81% |

**Table 5**. Performance evaluation metrics of used learning algorithms utilizing Gabor feature space.

| ML Models | Accuracy | Sensitivity | Precision | F1 Score | Specificity |
|---|---|---|---|---|---|
| kNN | 70% | 72% | 84% | 77% | 86% |
| RF | 78% | 68% | 85% | 69% | 88% |
| CNN | 78% | 72% | 82% | 77% | 84% |
| SVC | 94.8% | 93% | 93.9% | 94.5% | 96.6% |
| PNN | 81% | 70% | 89% | 78% | 91% |

**Table 6**. Performance evaluation metrics for various used classifiers utilizing GLRLM feature space.

| ML Models | Accuracy | Sensitivity | Precision | F1 Score | Specificity |
|---|---|---|---|---|---|
| kNN | 95% | 97% | 94% | 97% | 93% |
| RF | 95% | 98% | 94% | 97% | 92% |
| CNN | 97.8% | 97.7% | 97% | 96% | 97.9% |
| SVC | 98.06% | 97.4% | 97.44% | 96.20% | 98.7% |
| PNN | 93% | 97% | 90% | 94% | 89% |

**Table 7**. Performance evaluation metrics for various classifiers utilizing LBP feature space.

| Feature Space | $\chi^2$ (Chi-Square) Statistic | *P*-value ($\alpha = 0.05$) | Conclusion |
|---|---|---|---|
| CNN | 19.8 | 0.001 | Significant |
| DWT | 14.5 | 0.02 | Significant |
| FFT | 10.7 | 0.07 | Not significant |
| GRLM | 16.2 | 0.05 | Not significant |
| Gabor filter | 11.9 | 0.07 | Not significant |
| LBP | 19.3 | 0.003 | Significant |

**Table 8**. Chi-square and P-value of various feature representation schemes.

## GLRLM feature space

GLRLM-based information exhibited a robust correlation with SVC, obtaining an accuracy of 94.8% and 96.6% specificity are shown in Table 6. kNN yielded a balance between recall and precision, while CNN and RF successfully delivered correct positive predictions. In addition, PNN obtained 81% of accuracy which performed moderately. The performance variations indicate that GLRLM is appropriate for specific learning algorithms but not universally optimal. However, GLRLM representation methods extract texture information efficiently while its ability is limited to capture spatial relationships in images.

## LBP feature extraction

Local Binary Pattern (LBP) features show excellent outcomes which are shown in Table 7. SVC again obtained the highest success rate which is 98.06. The sensitivity and specificity of SVC are balanced. CNN yielded the second-highest accuracy of 97.8%, succeeded by RF at 95%. The balanced sensitivity, precision, and F1-score across learning algorithms validated the robustness of LBP features in brain tumor classification. LBP is effective in collecting local texture patterns and is also efficient and simple computationally.

In order to investigate the contribution of each feature representation scheme, statistical test chi-square and p-value are computed which is illustrated in Table 8. The results indicate that CNN, DWT, and LBP exhibit a significant impact ($p < 0.05$), thereby confirming their importance in the classification process. The results indicate that FFT, GRLM, and Gabor do not achieve statistical significance ($p > 0.05$), suggesting a minimal contribution to classification accuracy.

## Performance metrics of learning algorithms using large training dataset

The empirical results reveal that leaning algorithms have achieved the highest success rates on LBP feature representation methods. In order to further investigate and highlight the strength of LBP feature representation methods, the large dataset is expressed by the LBP method. In this dataset, the total number of MRI images are 7023 of which 5723 images are for training and 1311 images are for testing. The performance of learning algorithms on LBP feature space is shown in Table 9. The highest accuracy is achieved by CNN with 98.9% while SVC obtained 96.7% accuracy. The sensitivity and specificity of CNN are balanced. On the other hand, the performance of kNN, RF, and PNN are comparatively the same. The performance of learning algorithms on LBP feature space is still advanced compared to other learning algorithms in the case of large training datasets.

The AI explainable SHAP analysis (Fig. 2) revealed that the most influential LBP features correspond to regions with high texture contrast and edge definitions. These align with known MRI characteristics of brain tumors. For instance, uniform texture patterns with high SHAP scores were often associated with meningioma, while features indicative of texture heterogeneity were linked to gliomas. This correspondence affirms the model's clinical interpretability and the usefulness of the LBP-based features in capturing tumor characteristics that have medical significance. In Fig. 2 each row corresponds to a specific feature, where red points denote higher feature values and blue points indicate lower values. The SHAP values represented on the x-axis quantify the influence of features on model predictions. Positive SHAP values derive predictions toward the tumor, while negative values indicate healthy. Various feature subsets with different dimensions were evaluated, and the optimal selection of the highest-ranked features markedly improved the model's prediction performance.

In addition, Accuracy of individual classes of tumors on both datasets of CNN and SVC are listed in Table 10.

In this work, comparison analysis was also carried out, focused on interpretability, deployment feasibility, and computing efficiency in order to substantiate the advantages of our proposed deep learning model over the transformer-based model. Our proposed model performs excellently compared to the Swin Transformer baseline in terms of computational cost, exhibiting about twice the inference performance on an NVIDIA RTX 3080 GPU, about 65% fewer floating-point operations (FLOPs), and 50% less memory utilization. These enhancements make our proposed model a strong contender for deployment in resource-constrained environments, such as rural healthcare settings where restricted hardware capacity and power availability often prohibit the implementation of more intricate transformer architectures. Additionally, our proposed model CNN-SHAP approach, uses feature attribution maps to produce interpretable visualizations in order to recognize the interpretability and generalizability trade-offs. In medical deep learning, model transparency must guarantee diagnostic reliability without sacrificing clinical acceptability. This conflict has been highlighted in published research by emphasizing the necessity of explainability in high-stakes applications Doshi-Velez and Kim (2017) and Tjoa and Guan (2020). In order to make our method both efficient and useful for actual clinical applications, we strive for a balance between obtaining high performance on benchmark datasets and preserving transparency through SHAP-based post hoc explanations. The performance of the proposed model is evaluated using a stratified five-fold cross-validation test. The mean and standard deviation (mean ± SD) of each class are reported in Table 11. It demonstrates the consistency and reliability of the proposed model on both small and large datasets. In addition, 95% confidence intervals are also computed applying a Student's t-distribution (df = 4 and sample size $n = 5$) for each class and shown in Table 12. The narrow confidence intervals indicate high model stability and consistent performance.

## Discussion

A comprehensive performance evaluation indicated significant differences in the performance of various combinations of feature extraction methods and classifiers. The SVC exhibited consistently high accuracy and specificity, establishing it as a dependable model for the detection of brain tumors in MRI scans. This indicates that SVC demonstrates a high proficiency in differentiating between tumor and non-tumor cases while maintaining a low rate of false positives. The RF classifier demonstrated high precision, reflecting its effectiveness in producing accurate positive predictions and reducing the occurrence of misclassifications. The CNN model demonstrated significant performance improvements, particularly in minimizing false positive instances (Relu, softmax, Adam), thereby increasing diagnostic reliability. In the meantime, the kNN algorithm demonstrated high sensitivity, successfully identifying true positive cases. This emphasizes its ability to accurately identify the presence of tumors; however, additional optimization may be necessary to minimize false positives.

LBP space has shown significant effectiveness in capturing texture-based features, leading to high classification accuracy. This highlights the significance of selecting a suitable feature representation method to improve tumor detection. Additionally, this study emphasizes the critical importance of feature extraction methods in enhancing diagnostic accuracy within medical imaging applications.

| ML Models | Accuracy | Sensitivity | Precision | F1 Score | Specificity |
|---|---|---|---|---|---|
| kNN | 91% | 90% | 92% | 90% | 92% |
| RF | 91.6% | 91.2% | 92% | 91% | 92% |
| CNN | 98.9% | 98.6% | 99% | 98% | 99.2% |
| SVC | 96.7% | 95.4% | 96.2% | 96% | 98.% |
| PNN | 92% | 94% | 90% | 92% | 90% |

**Table 9.** Performance evaluation metrics for various classifiers utilizing LBP feature space on large dataset.

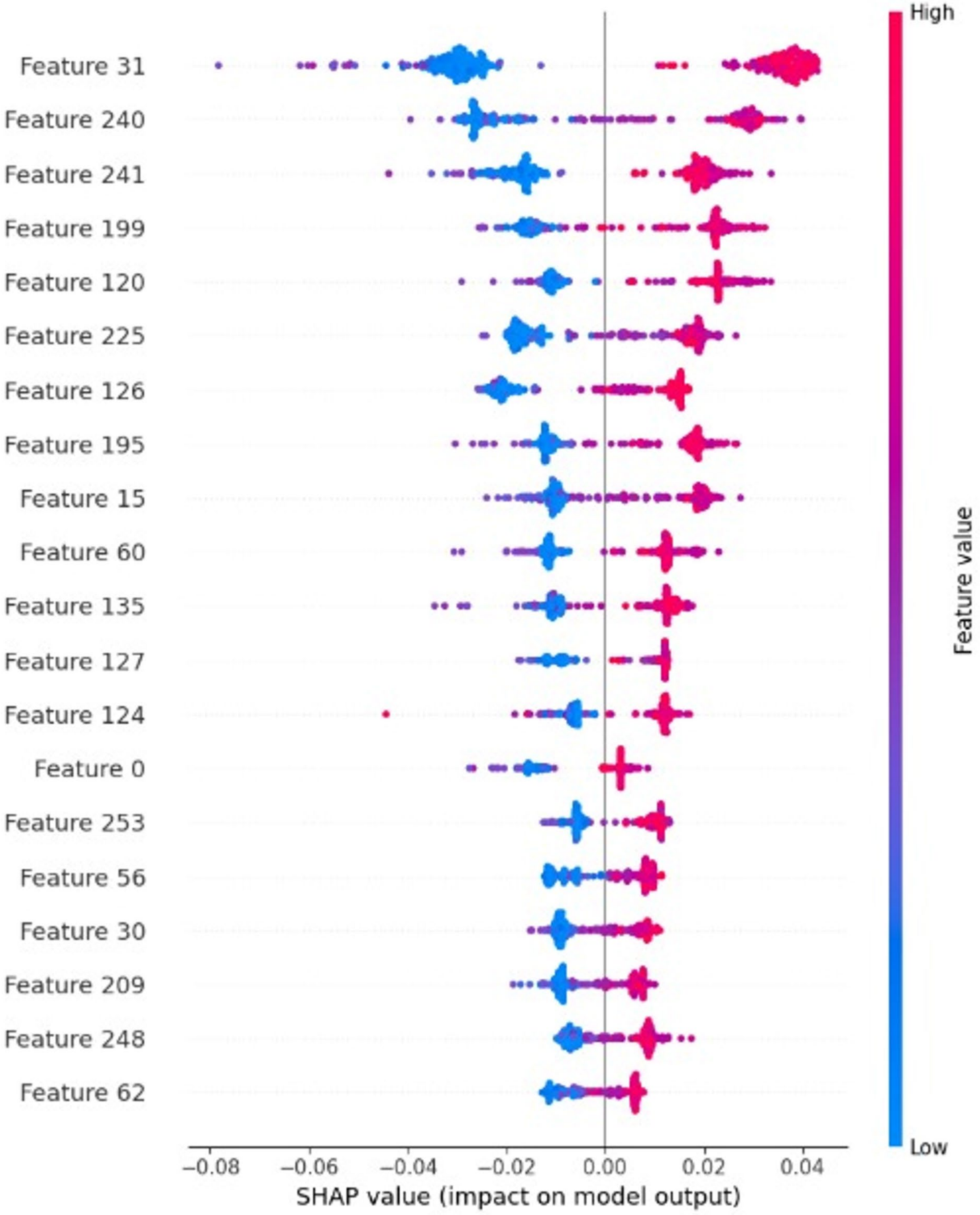

**Fig. 2**. SHAP interpretation of LBP features.

In the case of training dataset1, the prediction outcomes of SVC are reported the highest using all feature spaces. SVC obtained 98.06% accuracy on LBP feature space. The second highest success rates are obtained by CNN and are comparatively the same as SVC. CNN yielded 97.7% accuracy on LBP feature space. Further, evaluating the generalization and resilience of learning algorithms on the LBP feature representation method another large dataset training dataset2 is used. In the case of a large dataset expressed by the LBP feature representation method, CNN was identified as the highest-performing model, with a weighted accuracy of 98.9%, and SVC obtained 96.7% accuracy. In the case of the small dataset, the machine learning algorithm

| ML models | Tumors classes | | | Non-tumors |
|---|---|---|---|---|
| | Pituitary | Meningioma | Glioma | |
| Small dataset | | | | |
| CNN | 97.0% | 96.9% | 97.4% | 96% |
| SVC | 98.68% | 97.6% | 97.9% | 96.20% |
| Large dataset | | | | |
| CNN | 99.3% | 98.4% | 99% | 99.2% |
| SVC | 97.6% | 96.4% | 96.1% | 98.% |

**Table 10**. Accuracy of individual classes on both dataset of CNN and SVC.

| Dataset | Model | Pituitary (%) | Meningioma (%) | Glioma (%) | Non-Tumor (%) |
|---|---|---|---|---|---|
| Small | CNN | 97.0 ± 0.45 | 96.9 ± 0.51 | 97.4 ± 0.37 | 96.0 ± 0.59 |
| | SVC | 98.68 ± 0.32 | 97.6 ± 0.44 | 97.9 ± 0.41 | 96.20 ± 0.50 |
| Large | CNN | 99.3 ± 0.21 | 98.4 ± 0.26 | 99.0 ± 0.24 | 99.2 ± 0.19 |
| | SVC | 97.6 ± 0.39 | 96.4 ± 0.42 | 96.1 ± 0.47 | 98.0 ± 0.33 |

**Table 11**. Accuracy (mean ± SD) of individual classes on both small and large datasets.

| Dataset | Model | Pituitary (%) | Meningioma (%) | Glioma (%) | Non-Tumor (%) |
|---|---|---|---|---|---|
| Small | CNN | 96.44–97.56 | 96.27–97.53 | 96.94–97.86 | 95.27–96.73 |
| | SVC | 98.28–99.08 | 97.05–98.15 | 97.39–98.41 | 95.58–96.82 |
| Large | CNN | 99.04–99.56 | 98.08–98.72 | 98.70–99.30 | 98.96–99.44 |
| | SVC | 97.12–98.08 | 95.88–96.92 | 95.52–96.68 | 97.59–98.41 |

**Table 12**. 95% confidence intervals (CI) for accuracy.

SVC is well performed and the deep learning model CNN is very close in terms of accuracy. In contrast, the predictive outcomes of CNN in the case of large datasets are much better compared to SVC. The empirical results elaborate that as the number of images increased the performance of SVC reduced while the performance of CNN enhanced. So, machine learning algorithms are useful in the case of small datasets, in contrast, CNN performs well in the case of large datasets.

In conclusion, the results highlight the essential relationship between feature extraction methods and machine learning and deep learning classifiers in enhancing the accuracy of brain tumor detection. The findings demonstrate that LBP-based feature extraction yields strong feature representations. However, traditional methods such as DWT and Gabor filters also show competitive performance when utilized alongside effective classifiers like CNN, SVC, and RF. The insights presented have significant implications for enhancing the accuracy and reliability of automated diagnostic systems within the field of medical imaging.

CNN identifies spatial hierarchies of features, effectively capturing patterns including edges, textures, and intricate structures, thereby demonstrating high suitability for image-based tasks. The deep learning architecture enables the achievement of higher accuracy in large and complex datasets. The model demonstrates strong generalization capabilities across diverse image datasets through the acquisition of robust feature representations. The system effectively manages variations including rotation, scale, and intensity differences by utilizing convolutional layers and pooling mechanisms. SVC is dependent on pre-extracted features, indicating that its performance is significantly influenced by the quality and relevance of these features. The system does not acquire spatial hierarchies; instead, it identifies optimal decision boundaries for classification based on the input features. It demonstrates robust performance when supplied with well-extracted and relevant features. It may encounter difficulties when processing high-dimensional and raw image data unless extensive preprocessing is applied. The system exhibits increased sensitivity to variations in image data due to its lack of inherent spatial feature learning, resulting in reduced robustness when addressing diverse and complex image patterns. It is computationally less expensive in comparison to deep learning models. The method demonstrates effective performance on smaller datasets and exhibits ease of training; however, it may not achieve the same level of proficiency as CNN in processing highly complex image data. However, the dependence on extracted features indicates that performance may be suboptimal when encountering unseen or complex variations in medical images.

SVC demonstrates high effectiveness in structured classification tasks that utilize well-engineered features. However, CNN exhibits superior performance in medical image analysis, attributed to its capability to learn intricate spatial features, achieve robust generalization, and uphold elevated accuracy and specificity. Considering the intricacies involved in MRI brain tumor detection, CNN is identified as a more effective option for automated and high-precision medical imaging applications.

| Study | Methodology | Accuracy (%) | Sensitivity (%) | Precision (%) | F1-Score (%) | Specificity (%) |
|---|---|---|---|---|---|---|
| CNN-based model [75] | CNN as a feature extractor & classifier | 96.00 | - | 91.14 | - | - |
| Auto-encoder Network[76] | Auto-encoder + 2D CNN | 95.63, | 94 | - | - | - |
| kNN[77] | kNN classifier | 86.00 | - | - | - | - |
| CNN + classic Architecture[78] | CNN hybrid model | 96.00 | 95 | 95 | - | - |
| Gradient boosting[79] | Gradient Boosting | 92.40 | 94.40 | 85 | 89.5 | - |
| Brain tumor Classification[80] | Proposed classification method | 97.15 | 97 | 97 | 97 | - |
| BRATS 2018 Dataset[81] | Various approaches | 94.60 | - | - | - | - |
| Proposed model | LBP feature extraction + CNN | **98.9%** | **98.6%** | **99%** | **98%** | **99.2%** |

**Table 13**. Comparison of the proposed model with existing models.

## Comparative analysis of exiting models with propose model

This study distinguishes itself by employing multiple machine learning and deep learning algorithms in conjunction with diverse feature representation strategies, validating a comprehensive comparative analysis with existing research are reported in Table 13. The empirical results highlight the significance of our proposed approach in enhancing classification performance. For instance, a prior study[75] reported an accuracy of 93.3% using a CNN-based model. Similarly, an auto-encoder network and 2D CNN were proposed in[76], attaining 95.63% and 96.47% accuracy, respectively. The kNN classifier in[77] achieved an accuracy of 86%. A novel hybrid approach combining CNN with classical architectures was proposed in[78], yielding 96% accuracy along with 95% sensitivity and precision. Gradient Boosting, identified as the best-performing classifier in[79], achieved an accuracy of 92.4%, an F1-score of 89.5%, 94.4% sensitivity, and 85% precision. Another brain tumor classification study[80] attained 97.15% accuracy with 97% precision, recall, and F1-score. To further validate the robustness of the model, a new model was introduced BRATS 2018 obtained 94.6%, accuracy, respectively[81]. In contrast, our proposed model achieved 98.9% accuracy, 98.6% sensitivity, 99.0% precision, and 99.2% specificity.

The proposed model exhibits enhanced performance across all evaluated metrics compared to existing models. The integration of LBP feature space along with CNN for classification likely contributes to its high accuracy, sensitivity, precision, F1-score, and specificity. The improved performance of the model highlights the effectiveness of merging texture-based feature extraction with the deep learning algorithm CNN for MRI-based brain tumor classification. Consequently, it presents itself as a viable option for brain tumor classification tasks, especially in medical applications where minimizing false positives and false negatives is critical. However, further, the proposed model is validated on larger and more diverse datasets in order to confirm its generalizability and robustness.

## Conclusion

This study investigates the effectiveness of various feature representation techniques, including LBP, Gabor filters, DWT, FFT, CNN, and GLRLM. It also examines their application with different machine learning classifiers, such as CNN, RF, SVC, PNN, and CNN.

The experimental findings indicate that LBP, in conjunction with SVC and CNN, attains elevated specificity and accuracy, positioning it as a viable method for MRI-based tumor diagnosis. The statistical analysis employing chi-square and p-value tests validates the significant influence of LBP feature representation on brain tumor identification. In the analysis of smaller datasets, the CNN demonstrated an accuracy of 97.8%, whereas the SVC achieved an accuracy of 98.06%. Subsequent assessment using an expanded benchmark dataset indicated that the CNN model achieved the highest performance, recording an accuracy of 98.9%, while the SVC model followed with an accuracy of 96.7%. SVC demonstrates effectiveness in structured classification tasks that utilize well-engineered features. In contrast, CNN exhibits superior performance in medical imaging, attributable to its capacity to learn complex spatial features and generalize efficiently. In addition, The SHAP analysis was used to identify the most significant features in classification. The proposed model demonstrates superior performance compared to existing approaches across all critical metrics, such as accuracy, sensitivity, precision, F1-score, and specificity. The combination of texture-based feature extraction and deep learning, specifically CNN, improves the predictive capabilities of the model, rendering it particularly effective for the classification of brain tumors in MRI scans. This study highlights the necessity of employing advanced feature extraction and classification techniques to minimize false positives and false negatives in medical applications. In brain tumor classification challenges, this study shows that traditional texture-based feature extraction techniques can perform competitively when paired with shallow classifiers. These techniques are useful substitutes or supplements to deep learning because they offer improved interpretability, lower computing overhead, and increased robustness across datasets, especially in clinical settings with low data and high interpretability.

Future research should concentrate on validating the proposed model using larger and more diverse datasets to enhance confirmation of its robustness and generalizability. Future research will expand the analysis to include transformer-based architectures and self-supervised models like DINO and CLIP, especially as larger annotated medical datasets become available, even though this study concentrates on classical and CNN-based models because of their effectiveness and suitability for smaller datasets.

## Data availability

The datasets used and/or analysed during the current study available from the corresponding author on reasonable request.

## Acknowledgements
The Researchers would like to thank the Deanship of Graduate Studies and Scientific Research at Qassim University for financial support (QU-APC-2025).

## Author contributions
Conceptualization: Maqsood Hayat, Fawaz Khaled Alarfaj, Salem Alkhalaf, Fahad AlturiseData curation: Maqsood Hayat, Salem AlkhalafFunding acquisition: Salem Alkhalaf, Fahad Alturise.Investigation: Asif Rah-man, Maqsood Hayat, Fawaz Khaled Alarfaj, Salem Alkhalaf, Fahad AlturiseMethodology: Asif Rahman, Nadeem Iqbal, Maqsood Hayat, Fawaz Khaled Alarfaj, Salem Alkhalaf, Fahad Alturise.Project administration: Maqsood Hayat, Fawaz Khaled Alarfaj, Salem Alkhalaf, Fahad Alturise.Resources: Salem Alkhalaf.Software: Maqsood HayatSupervision: Maqsood Hayat, Salem AlkhalafValidation: Aisf Rahman, Nadeem Iqbal, Maqsood Hayat, Fawaz Khaled Alarfaj. Salem Alkhalaf, Fahad AlturiseVisualization: Asif Rahman, Maqsood Hayat, Fawaz Khaled Alarfaj Writing – original draft: Asif Rahman, Nadeem Iqbal, Maqsood Hayat, Fawaz Khaled Alarfaj. Salem Alkhalaf, Fahad Alturise Writing – review & editing: Asif Rahman, Maqsood Hayat, Fawaz Khaled Alarfaj. Salem Alkhalaf, Fahad Alturise.

## Declarations

### Competing interests
The authors declare no competing interests.

### Additional information
**Correspondence** and requests for materials should be addressed to M.H. or S.A.